\documentclass[letterpaper, 10 pt, conference]{packages/ieeeconf}

\IEEEoverridecommandlockouts
\usepackage{multirow}
\usepackage{graphics} 
\usepackage{epsfig} 
\usepackage{times} 
\usepackage{amssymb}  
\usepackage{mathtools}
\usepackage{packages/commath}

\usepackage{amsthm}
\usepackage{graphicx}
\usepackage{subfloat}
\usepackage{subcaption}

\usepackage{array}
\newcolumntype{M}[1]{>{\centering\arraybackslash}m{#1}}

\usepackage{colortbl}
\usepackage{tabulary}

\usepackage{amsmath,amsfonts} 

\usepackage{amsmath,amssymb,amsfonts}
\usepackage{graphicx}
\DeclareGraphicsExtensions{.eps,.pdf,.png,.jpg,.JPG,.gif}
\usepackage{subfloat}
\usepackage{algorithm}
\usepackage{algorithmic}
\usepackage{bm}
\usepackage{xcolor}
\usepackage[colorinlistoftodos]{todonotes}

\usepackage{array}
\newcolumntype{M}[1]{>{\centering\arraybackslash}m{#1}}

\usepackage{array}
\usepackage{booktabs}
\newcommand{\PreserveBackslash}[1]{\let\temp=\\#1\let\\=\temp}
\newcolumntype{C}[1]{>{\PreserveBackslash\centering}m{#1}}
\newcolumntype{R}[1]{>{\PreserveBackslash\raggedleft}p{#1}}
\newcolumntype{L}[1]{>{\PreserveBackslash\raggedright}p{#1}}

\usepackage{xcolor}

\title{\huge \bf
		LIC-Fusion: LiDAR-Inertial-Camera Odometry}

\author{Xingxing Zuo$^\ast$, Patrick Geneva$^\dagger{}^\dagger$, Woosik Lee$^\dagger$, Yong Liu$^\ast$, and Guoquan Huang$^\dagger$
    \thanks{This work is supported in part by the National Natural Science Foundation of China under Grant U1509210, 61836015. P. Geneva was also  partially supported by the Delaware Space Grant College and Fellowship Program (NASA Grant NNX15AI19H).}%
	\thanks{$^\ast$The authors are with the Institute of Cyber-System and Control, Zhejiang University, Hangzhou, China. (Y. Liu is the corresponding
author).
		Email: {\tt\small xingxingzuo@zju.edu.cn,  yongliu@iipc.zju.edu.cn} }%
	\thanks{$^\dagger$The authors are with the Department of Mechanical Engineering, University of Delaware, Newark, DE 19716, USA.
		Email: {\tt\small \{ghuang,woosik\}@udel.edu}}%
	\thanks{$^\dagger{}^\dagger$The author is with the Department of Computer and Information Sciences, University of Delaware, Newark, DE 19716, USA.
		Email: {\tt\small pgeneva@udel.edu}}%
}

\begin{document}
	
\maketitle

\begin{abstract}
	
	This paper presents a tightly-coupled multi-sensor fusion algorithm termed LiDAR-inertial-camera fusion (LIC-Fusion),
	which efficiently fuses  IMU measurements, sparse visual features, and extracted LiDAR points.
	In particular, the proposed LIC-Fusion performs online spatial and temporal sensor calibration between all three asynchronous sensors,
	in order to compensate for possible calibration variations.
	The key contribution is the  optimal (up to linearization errors) multi-modal sensor fusion of detected and tracked sparse edge/surf feature points from LiDAR scans within an efficient MSCKF-based framework, alongside  sparse visual feature observations and IMU readings.
	%
	%
	We perform extensive experiments in both indoor and outdoor environments,
	showing that the proposed LIC-Fusion outperforms the state-of-the-art visual-inertial odometry (VIO) and LiDAR odometry methods in terms of estimation accuracy and robustness to aggressive motions.
	
\end{abstract}

\section{Introduction and Related Work}

%

It is essential to be able to accurately track the 3D motion of autonomous vehicles and mobile perception systems. 
One popular solution is inertial navigation systems (INS) aided with a monocular camera, 
which has recently attracted significant attention \cite{Mourikis2007ICRA,qin2018vins,mur2017visual,Huang2015ISRR,leutenegger2013keyframe,Huai2018IROS},
in part because of their complimentary sensing modalities, low cost, and small size.
However, cameras are limited by lighting conditions and cannot provide high-quality information in low-light or nighttime conditions.
In contrast, 3D LiDAR sensors can provide more robust and accurate range measurements regardless of lighting condition, and are therefore popular for robot localization and mapping \cite{zhang2014loam, park2017probabilistic, legoloam2018, behley2018efficient}.
3D LiDARs suffer from point cloud sparsity, high cost, and lower collection rates as compared to cameras.
LiDARs are still expensive as of today, limiting their widespread adoptions, but are expected to have dramatic cost reduction in coming years due to emerging new technology~\cite{lidar-dev}.
Inertial measurement units (IMUs) measure local angular velocity and linear acceleration and can provide large amount of information in dynamic trajectories but exhibit large drift due to noises if not fused with other information.
In this work, we focus on LiDAR-inertial-camera odometry (LIC) which offers the ``best'' of each sensor modality to provide a fast and robust 3D motion tracking solution in all scenarios.

\begin{figure}
	\centering
	\includegraphics[width=\columnwidth]{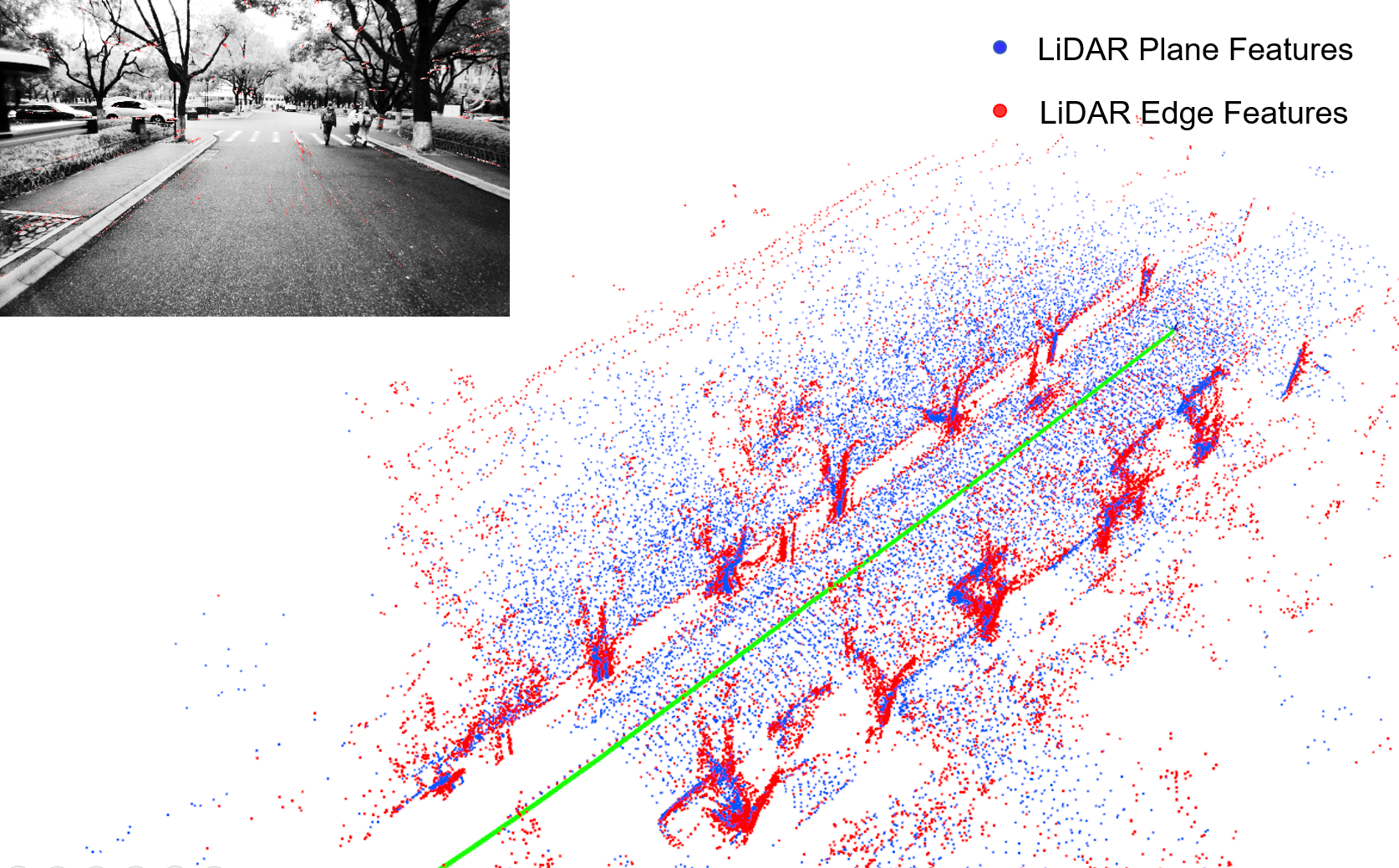} 
	\caption{The proposed LIC-Fusion fuses both sparse visual features tracked in images and  LiDAR features extracted in point clouds. The LiDAR points in red and blue are  the edge and plane features, respectively. Estimated trajectory is marked in green.
	}
	\label{fig: overview}
\end{figure}

Fusing these multi-modal measurements, in particular, from camera and LiDAR, is often addressed within a SLAM framework~\cite{graeter2018limo}.
For example,  Zhang, Kaess, and Singh~\cite{zhang2014real} associated depth information from LiDAR to visual camera features, resulting in what can be considered as a RGBD system with augmented LiDAR depth.
Later, Zhang and
Singh~\cite{zhang2015visual} developed a general framework for combining visual odometry (VO) and LiDAR odometry
which uses high-frequency visual odometry to estimate the overall ego-motion while lower-rate LiDAR odometry, which matches scans to the map and refines the VO estimates.
Shin, Park, and Kim~\cite{shin2018direct} have used the depth from LiDAR in a direct visual SLAM method, where photometric errors were minimized in an iterative way. 
Similarly, in~\cite{graeter2018limo}  LiDAR was leveraged for augmenting depth to visual features by fitting local planes, which was shown to perform well in autonomous driving scenarios.

Recently, Zhang and Singh~\cite{zhang2018laser}  developed a laser visual-inertial odometry and mapping system which employed a sequential multi-layer processing pipeline and consists of three main components: IMU prediction, visual-inertial odometry, and scan matching refinement.
Specifically, IMU measurements are used for prediction,
and the visual-inertial subsystem performs iterative minimization of a joint cost function of the IMU preintegration and visual feature re-projection error.
Then, LiDAR scan matching is performed via iterative closet point (ICP) to further refine the prior pose estimates  from the VIO module.
%
%
Note that both the iterative optimization and ICP  require sophisticated pipelines and parallel processing to allow for realtime performance.
Note also that 
this essentially is a loosely-coupled fusion approach because 
only the pose estimation results from the VIO is fed into the LiDAR scan matching subsystem 
and the scan matching cannot directly process the raw visual-inertial measurements, losing correlation information between LiDAR and VIO.

In this paper, we develop a fast, {tightly-coupled}, single-thread, LiDAR-inertial-camera (LIC) odometry algorithm with online spatial and temporal multi-sensor calibration within the computationally-efficient multi-state constraint kalman filter (MSCKF) framework~\cite{Mourikis2007ICRA}.
%
%
%
The main contributions of this work are the following:
\begin{itemize}
	\item We develop a tightly-coupled LIC odometry (termed LIC-Fusion), which enables efficient 6DOF pose estmation with online spatial and temporal calibration.
	The proposed LIC-Fusion  efficiently combines  IMU measurements, sparse visual features, and two different sparse LiDAR features (see Figure~\ref{fig: overview}) within the MSCKF framework.
	The dependence of the calibrated extrinsic parameters and estimated poses on measurements is explicitly modeled and analytically derived.
	
	
	\item We perform extensive experimental validations of the proposed approach on real-world experiments including indoor and outdoor environments, showing that the proposed LIC-Fusion is more accurate and more robust than state-of-the-art methods.
\end{itemize}
%


\section{The Proposed LIC-Fusion}

In this section, we present in detail the proposed LIC-Fusion odometry that tightly fuses LiDAR, inertial, and camera measurements within the MSCKF~\cite{Mourikis2007ICRA} framework.

\subsection{State Vector}
The state vector of the proposed method includes the IMU state $\mathbf{x}_{I}$ at time $k$, the extrinsics between IMU and camera $\mathbf{x}_{calib\_C}$, the extrinsics between IMU and LiDAR $\mathbf{x}_{calib\_L}$, a sliding window of clones, including local IMU clones at the past $m$ image times $\mathbf{x}_{C}$ and at the past $n$ LiDAR scan times $\mathbf{x}_{L}$.
The total state vector is:
\begin{align}
	\label{eq:state vector}
	\mathbf{x} & = 
	\begin{bmatrix}
		\mathbf{x}^{\top}_{I} & \mathbf{x}^{\top}_{calib\_C} & \mathbf{x}^{\top}_{calib\_L} & \mathbf{x}^{\top}_{C}  & \mathbf{x}^{\top}_{L}
	\end{bmatrix}^{
		\top}
\end{align}
where
\begin{align}
	\mathbf{x}_{I} & = 
	\begin{bmatrix}
		{}^{I_k}_G\bar{q}^{\top} & \mathbf{b}^{\top}_{g} & {}^G\mathbf{v}^{\top}_{I_k} & \mathbf{b}^{\top}_{a} & {}^G\mathbf{p}^{\top}_{I_k}
	\end{bmatrix}^{\top}  \\
	\mathbf{x}_{calib\_C} & = 
	\begin{bmatrix}
		{}^C_I\bar{q}^{\top} &  {}^C\mathbf{p}^{\top}_{I} & t_{dC}
	\end{bmatrix}^{\top} \\
	\mathbf{x}_{calib\_L} & = 
	\begin{bmatrix}
		{}^L_I\bar{q}^{\top} &  {}^L\mathbf{p}^{\top}_{I} & t_{dL}
	\end{bmatrix}^{\top}\\
	\mathbf{x}_{C} &=
	\begin{bmatrix}
		{}^{I_{a_1}}_G\bar{q}^{\top}  &   {}^G\mathbf{p}_{I_{a_1}}^{\top} &\!   \cdots  \!&  {}^{I_{a_m}}_G\bar{q}^{\top} &  {}^G\mathbf{p}_{I_{a_m}}^{\top} 
	\end{bmatrix}^\top \\
	\mathbf{x}_{L} &=
	\begin{bmatrix}
		{}^{I_{b_1}}_G\bar{q}^{\top}  &   {}^G\mathbf{p}_{I_{b_1}}^{\top} &\!   \cdots  \!&  {}^{I_{b_n}}_G\bar{q}^{\top} &  {}^G\mathbf{p}_{I_{b_n}}^{\top} 
	\end{bmatrix}^\top
\end{align}
${}^{I_k}_G\bar{q}$ is the JPL quaternion~\cite{Trawny2005_Q_TR} corresponding to the 3D rotation matrix ${}^{I_k}_G\mathbf{R}$, which denotes the rotation from the global  frame of reference $\{G\}$ to the local frame $\{I_k\}$ of IMU at time instant $t_k$. ${}^{G}\mathbf{v}_{I_k}$ and ${}^G\mathbf{p}_{I_k}$ represent the IMU velocity and position at time instant $t_k$ in the global frame, respectively.
$\mathbf{b}_{g}$ and $\mathbf{b}_{a}$ are the biases of gyroscope and accelerometer. 
${}^C_I\bar{q}$ and ${}^C\mathbf{p}_I$ represent the rigid-body transformation between the camera sensor frame $\{C\}$ and the IMU frame $\{I\}$.
Analogously, ${}^L_I\bar{q}$ and ${}^L\mathbf{p}_I$ is the 3D rigid transformation between the LiDAR and IMU frames.

We also co-estimate the time offsets between the exteroceptive sensors and the IMU, which commonly exist in low-cost devices due to sensor latency, clock skew, or data transmission delays.
Taking the IMU time to be the ``true'' base clock, we model that both the camera and LiDAR as having an offset $t_{dC}$ and $ t_{dL}$ which can correct the measurement time as follows:
\begin{align}
	t_I &= t_C + t_{dC} \\
	t_I &= t_L + t_{dL}
\end{align}
where $t_C$ and $t_L$ are the reported time in the camera and LiDAR clock respectively.
We refer the reader to~\cite{li2014online} for further details.
%
%
%
%
%

In the paper, we define that the true value of the state as $\mathbf{x} $, estimated
value as $ \hat{\mathbf{x}}$ , and corresponding error state $\delta \mathbf { x }$, is related by the following generalized update operation:
\begin{align}
	\mathbf { x }  = \hat { \mathbf { x } } \boxplus \delta \mathbf { x } 
\end{align}
The operation $\boxplus$ for a state  $\mathbf{v}$ in the vector space is simply the Euclidean addition, i.e., $\mathbf{v}= \hat{\mathbf{v}} + \delta \mathbf{v}$,
while for quaternion, it is given by:
\begin{align}
	\bar q \approx \begin{bmatrix} \frac{1}{2}\delta \bm \theta \\ 1 \end{bmatrix} \otimes \hat{\bar q}
\end{align}
where $\otimes$ denotes the JPL quaternion multiplication~\cite{Trawny2005_Q_TR}.

\subsection{IMU Propagation}

The IMU provides angular rate and linear accelerations measurements which we model with the following continuous-time kinematics \cite{Trawny2005_Q_TR}:
\begin{align}
	{}^{I_k}_G\dot{\bar{q}}(t) & = \frac{1}{2}\boldsymbol{\Omega}\left({}^{I_k}\boldsymbol{\omega}(t)\right){}^{I_k}_G\bar{q}(t)  \\
	{}^G\dot{\mathbf{p}}_{I_k}(t) & = {}^G\mathbf{v}_{I_k}(t)  \\ 
	{}^G\dot{\mathbf{v}}_{I_k}(t) &=  {}^{I_k}_G\mathbf{R}(t)^\top{}^{I_k}\mathbf{a}(t) + {}^G\mathbf{g}\\
	\dot{\mathbf{b}}_{g}(t) & = \mathbf{n}_{wg} \\
	\dot{\mathbf{b}}_{a}(t) &= \mathbf{n}_{wa}
	\label{eq:system model}
\end{align}
where $\mathbf{\Omega}(\bm \omega) = \bigl[ \begin{smallmatrix}
- \lfloor \bm \omega \rfloor & \bm \omega \\ - \bm \omega ^\top & \bm 0
\end{smallmatrix} \bigr]$, $\lfloor \cdot \rfloor$ is the skew symmetric matrix, ${}^{I_k}\boldsymbol{\omega}$ and ${}^{I_k}\mathbf{a}$ represent the true angular velocity and linear acceleration in the local IMU frame, and ${}^{G}\mathbf{g}$ denotes the gravitational acceleration in the global frame.
The gyroscope and accelerometer biases $\mathbf{b}_{g}$ and $\mathbf{b}_{a}$ are modeled as random walks, which are driven by the white Gaussian noises $\mathbf{n}_{wg}$ and $\mathbf{n}_{wa}$, respectively.
This continuous-time system can then be integrated and linearized to propagate the state covariance matrix forward in time \cite{Mourikis2007ICRA}.


\subsection{State Augmentation}

When the system receives a new image or LiDAR scan, the IMU state will propagate forward to that time instant, and the propagated inertial state is cloned into either the $\mathbf{x}_{C}$ or $\mathbf{x}_{L}$ state vectors.
In order to calibrate the time offsets between different sensors, we will propagate up to IMU time $\hat{t}_{I_k}$, which is the current best estimate of the measurement collection time in the IMU clock.
For example, if a new LiDAR scan is received with timestamp $t_{L_k}$, we will propagate up to 
$\hat{t}_{I_k} = t_{L_k}+\hat{t}_{dL}$, and augment the state vector $\mathbf{x}_{L}$ to include this new cloned state estimate:
\begin{align}
	\hat{\mathbf{x}}_{L_k}(\hat{t}_{I_k}) = \begin{bmatrix}  {{}^{ I _ { k } }_G \hat{\bar{ q }}}(\hat{t}_{I_k})^ { \top }   & {{}^ { G }\hat{\mathbf { p }} _ { I _ { k } }}(\hat{t}_{I_k}) ^ { \top } \end{bmatrix}^ { \top }
\end{align}
We also augment the covariance matrix as:
\begin{align}
	\mathbf{P}(\hat{t}_{I_k}) \gets \begin{bmatrix}
		\mathbf{P}(\hat{t}_{I_k}) & \mathbf{P}(\hat{t}_{I_k}) \mathbf{J}_{I_k}(\hat{t}_{I_k})^\top \\
		\mathbf{J}_{I_k}(\hat{t}_{I_k}) \mathbf{P}(\hat{t}_{I_k}) & \mathbf{J}_{I_k}(\hat{t}_{I_k}) \mathbf{P}(\hat{t}_{I_k}) \mathbf{J}_{I_k}(\hat{t}_{I_k})^\top
	\end{bmatrix}
\end{align}
where $ \mathbf{J}_{I_k}(\hat{t}_{I_k})$ is the Jacobian of the new cloned $\hat{\mathbf{x}}_{L_k}(\hat{t}_{I_k})$ with respect to the current state~\eqref{eq:state vector}:
\begin{align}
	\mathbf{J}_{I_k}(\hat{t}_{I_k}) =   \frac{\partial \delta\mathbf{x}_{L_k}(\hat{t}_{I_k}) }{\partial \delta\mathbf{x}} &= \begin{bmatrix}
		\mathbf{J}_{I} &\mathbf{J}_{calib\_C}   &\mathbf{J}_{calib\_L} &\mathbf{J}_{C}  &\mathbf{J}_{L} 
	\end{bmatrix}\notag
\end{align}
In the above expression, $\mathbf{J}_{I}$ is the Jacobian with respect to the IMU state $\mathbf{x}_I$, given by:
\begin{align}
	\mathbf{J}_{I} = \begin{bmatrix}
		\mathbf{I}_{3 \times 3} & \mathbf{0}_{3\times 9} & \mathbf{0}_{3\times 3} \\
		\mathbf{0}_{3\times 3} & \mathbf{0}_{3\times 9} & \mathbf{I}_{3 \times 3} 
	\end{bmatrix}
\end{align} 
$\mathbf{J}_{calib\_L}$ is the Jacobian with respect to the extrinsics (including time offset) between IMU and LiDAR:
\begin{align}
	\mathbf{J}_{calib\_L} = \begin{bmatrix}
		\mathbf{0}_{6 \times 6} & \mathbf{J}_{t_{dL}}\end{bmatrix},~\mathbf{J}_{t_{dL}} = \begin{bmatrix}
		{}^{I_k}\hat{\boldsymbol{\omega}}^\top & {}^G\hat{\mathbf{v}}_{I_k}^\top
	\end{bmatrix}^\top
\end{align}
and ${}^{I_k}\hat{\boldsymbol{\omega}}$ denotes the local angular velocity of IMU at time $\hat{t}_{I_k}$, and ${}^G\hat{\mathbf{v}}_{I_k}$ is the global linear velocity of IMU at time $\hat{t}_{I_k}$.
%
Similarly, $\mathbf{J}_{calib\_C}$,  $\mathbf{J}_{C}$,  $\mathbf{J}_{L}$ are the Jacobian with respect to extrinsics between IMU and camera, clones at camera time, clones at LiDAR time, respectively, which should be zero in this case.
It is important to note that 
the dependence of the new cloned IMU state corresponding to the LiDAR measurement on $t_{dL}$ is modeled via the IMU kinematics 
and thus allows our measurement models (see  Section~\ref{sec:lidar-meas}) 
to be directly a function of the clones which are at the ``true'' measurement time in the IMU clock frame.
This LiDAR cloning procedure is analogous to the procedure used for when a new camera measurement occurs.
%

\subsection{Measurement Models}

\subsubsection{LiDAR Feature Measurement}
\label{sec:lidar-meas}

To limit the required computational cost, we wish to select a sparse set of high quality features from the raw LiDAR scan for state estimation.
In analogy to \cite{zhang2014loam}, we extract high and low curvature sections of LiDAR scan rings which correspond to edge and planar surf features respectively (see Figure \ref{fig: overview}).
We track the extracted edge and surf features in the current LiDAR scan back to the previous scan
by projecting and finding the closest corresponding features using KD-tree for fast indexing~\cite{de1997computational}.
%
For example, we project one feature point ${}^{L_{l+1}} \mathbf{p}_{fi}$ in the LiDAR scan $\{L_{l+1}\}$ to  $\{L_{l}\}$, 
the projected point is denoted as ${}^{L_{l}} \mathbf{p}_{fi}$:
\begin{align}
	{}^{L_{l}}\mathbf{p}_{fi} = 	{}^{L_l}_{L_{l+1}}\mathbf{R} {}^{L_{l+1}}\mathbf{p}_{fi} + {}^{L_l}\mathbf{p}_{L_{l+1}}
\end{align}
where ${}^{L_l}_{L_{l+1}} \mathbf{R}$ and ${}^{L_l}\mathbf{p}_{L_{l+1}}$ are the relative rotation and translation between two LiDAR frames, which can be computed from the states in the state vector:
\begin{align}
	{}^{L_l}_{L_{l+1}} \mathbf{R} & = {}^{L}_{I}\mathbf{R}  {}^{I_{l}}_G \mathbf{R} \left({}^{L}_{I}\mathbf{R} {}^{I_{l+1}}_G\mathbf{R}\right)^\top  \\
	{}^{L_l}\mathbf{p}_{L_{l+1}} & = {}^{L}_{I}\mathbf{R} {}^{I_{l}}_G \mathbf{R} \left( {}^{G}\mathbf{p}_{I_{l+1}} - {}^{G}\mathbf{p}_{I_{l}} + {}^{I_{l+1}}_{G}\mathbf{R}^\top   {}^{I}\mathbf{p}_{L}  \right) + {}^{L}\mathbf{p}_{I}  \\
	{}^{I}\mathbf{p}_{L} &=  -{}^{L}_{I}\mathbf{R}^\top {}^{L}\mathbf{p}_{I}
\end{align}

After this tracking,  we would find two  edge features in the old scan, ${}^{L_{l}} \mathbf{p}_{fj}, {}^{L_{l}} \mathbf{p}_{fk}$, 
corresponding to the projected edge feature ${}^{L_{l}} \mathbf{p}_{fi}$.
We assume they are sampled from the same physical edge as ${}^{L_{l}} \mathbf{p}_{fi}$. 
If the closest edge feature ${}^{L_{l}} \mathbf{p}_{fj}$  is on the $r$-th scan ring, 
then the second nearest edge feature ${}^{L_{l}} \mathbf{p}_{fk}$ should be on the immediate neighboring ring $r-1$ or $r+1$.
As a result, 
the measurement residual of the edge feature ${}^{L_{l+1}} \mathbf{p}_{fi}$ is the distance between its projected feature point ${}^{L_{l}} \mathbf{p}_{fi}$ and the straight line formed  by  ${}^{L_{l}} \mathbf{p}_{fj}$ and ${}^{L_{l}} \mathbf{p}_{fk}$:
\begin{align}
	r({}^{L_{l+1}} \mathbf{p}_{fi}) = \frac{\left\| \left( {}^{L_{l}}\mathbf{p}_{fi} - {}^{L_{l}}\mathbf{p}_{fj}\right) \times \left( {}^{L_{l}}\mathbf{p}_{fi} - {}^{L_{l}}\mathbf{p}_{fk}\right) \right\|_{2}}     {\left \|  {}^{L_{l}}\mathbf{p}_{fj} - {}^{L_{l}}\mathbf{p}_{fk} \right \|_{2} }
\end{align}
where ${\left \| \cdot  \right \|_{2}} $ is the Euclidean norm and $\times$ denotes the cross product of two vector.

We linearize the above distance measurement of edge features at the current state estimate:
\begin{align}
	r({}^{L_{l+1}} \mathbf{p}_{fi}) &= h(\mathbf{x}) + n_r \notag \\
	&= h(\hat{\mathbf{x}}) + \mathbf{H}_{\mathbf x}  \delta\mathbf{x} + n_r
	\label{eq: lidar residual}
\end{align}
where $\mathbf{H}_\mathbf{x}$ is the Jacobian of the distance with respect to the states in the state vector and $n_r$ is modeled as white Gaussian with variance $C_r$.
The non-zero elements in $\mathbf{H}_\mathbf{x}$ are only related to the cloned poses ${}^{I_l}_G\bar{q}, {}^{G}\mathbf{p}_{I_{l}}$ and ${}^{I_{l+1}}_G\bar{q}, {}^{G}\mathbf{p}_{I_{l+1}}$ along with the rigid calibration between the IMU and LiDAR ${}^{L}_I\bar{q}, {}^L\mathbf{p}_{I}$.
Thus we have:
\begin{align}
	\mathbf{H}_\mathbf{x} = \frac{ \partial \delta r({}^{L_{l+1}} \mathbf{p}_{fi})}     {\partial {}^{L_{l}} \delta\mathbf{p}_{fi}} \frac{{\partial {}^{L_{l}} \delta\mathbf{p}_{fi}}}{\partial \delta \mathbf{x}}
\end{align}
the non-zero elements in $\frac{{\partial {}^{L_{l}} \delta\mathbf{p}_{fi}}}{\partial \delta \mathbf{x}}$ are computed as:
{
	\allowdisplaybreaks
	\begin{align}
		\frac{{\partial {}^{L_{l}} \delta\mathbf{p}_{fi}}} {\partial {}^{I_{l}}_G \delta \bm \theta } &=   
		{}^{L}_I \hat{\mathbf{R}} \lfloor {}^{I_l}_G \hat{\mathbf{R}} {}^{I_{l+1}}_G \hat{\mathbf{R}}^\top {}^{L}_I \hat{\mathbf{R}} {}^{L_{l+1}}\mathbf{p}_{fi} \rfloor \notag\\ &~~~~+ {}^{L}_I \hat{\mathbf{R}} \lfloor {}^{I_l}_G \hat{\mathbf{R}} ( {}^{G}\hat{\mathbf{p}}_{I_{l+1}} - {}^{G}\hat{\mathbf{p}}_{I_{l}} + {}^{I_{l+1}}_{G}\hat{\mathbf{R}}^\top   {}^{I}\hat{\mathbf{p}}_{L}  )\rfloor
		\notag\\
		\frac{{\partial {}^{L_{l}} \delta\mathbf{p}_{fi}}} {\partial {}^{G}\delta \hat{\mathbf{p}}_{I_{l}} } &= -{}^{L}_I \hat{\mathbf{R}} {}^{I_l}_G \hat{\mathbf{R}}
		\notag\\		
		\frac{{\partial {}^{L_{l}} \delta\mathbf{p}_{fi}}} {\partial {}^{I_{l+1}}_G \delta \bm \theta } &=-{}^{L}_I \hat{\mathbf{R}} {}^{I_l}_G \hat{\mathbf{R}} {}^{I_{l+1}}_G \hat{\mathbf{R}}^\top \lfloor {}^{L}_I \hat{\mathbf{R}}^\top {}^{L_{l+1}}\mathbf{p}_{fi} +  {}^{I}\hat{\mathbf{p}}_{L} \rfloor\notag\\
		\frac{{\partial {}^{L_{l}} \delta\mathbf{p}_{fi}}} {\partial {}^{G}\delta \hat{\mathbf{p}}_{I_{l+1}} } &= {}^{L}_I \hat{\mathbf{R}} {}^{I_l}_G \hat{\mathbf{R}}\notag\\
		\frac{{\partial {}^{L_{l}} \delta\mathbf{p}_{fi}}} {\partial {}^{L}_I \delta \bm \theta } &= \lfloor {}^{L_l}_{L_{l+1}} \hat{\mathbf{R}} ( {}^{L_{l+1}}\mathbf{p}_{fi} - {}^L\hat{\mathbf{p}}_{I}) \rfloor \notag\\
		&~~~~ - {}^{L_l}_{L_{l+1}} \hat{\mathbf{R}}\lfloor {}^{L_{l+1}}\mathbf{p}_{fi} - {}^L\hat{\mathbf{p}}_{I} \rfloor \notag\\
		\frac{{\partial {}^{L_{l}} \delta\mathbf{p}_{fi}}} {\partial {}^{L}\delta \hat{\mathbf{p}}_{I} } &= - {}^{L_l}_{L_{l+1}} \hat{\mathbf{R}} + \mathbf{I}_{3\times 3} \notag
	\end{align}
}
In order to perform EKF update, we need to know the explicit covariance $C_r$ of the distance measurement.
As this measurement is not directly obtained from the LiDAR sensor, we propagate the covariance of raw measurements (point) in LiDAR scan $C_r$.
Assuming the covariance of point ${}^{L_{l+1}} \mathbf{p}_{fi}, {}^{L_{l}} \mathbf{p}_{fj}, {}^{L_{l}} \mathbf{p}_{fk}$ are $\mathbf{C}_i, \mathbf{C}_j,\mathbf{C}_k$ respectively, $C_r$ can be computed as:
\begin{align}
	C_r &= \sum_{x=i,j,k}^{} \mathbf{J}_x \mathbf{C}_x \mathbf{J}_x^\top,~~
	\mathbf{J}_i = \frac{ \partial \delta r({}^{L_{l+1}} \mathbf{p}_{fi})}     {\partial {}^{L_{l+1}} \delta\mathbf{p}_{fi}} \notag\\ 
	\mathbf{J}_j &= \frac{ \partial \delta r({}^{L_{l+1}} \mathbf{p}_{fi})}     {\partial {}^{L_{l}} \delta\mathbf{p}_{fj}},~~
	\mathbf{J}_k = \frac{ \partial \delta r({}^{L_{l+1}} \mathbf{p}_{fi})}     {\partial {}^{L_{l}} \delta\mathbf{p}_{fk}}
\end{align}
%
%
We perform simple probabilistic outlier rejection based on the Mahalanobis distance:
\begin{align}
	r_{m} & = r({}^{L_{l+1}} \mathbf{p}_{fi})^\top \left(
	\mathbf{H}_\mathbf{x} 
	\mathbf{P}_\mathbf{x} \mathbf{H}_\mathbf{x}^\top 
	+ 
	C_r
	\right)^{-1} r({}^{L_{l+1}} \mathbf{p}_{fi}) \notag
\end{align}
where $\mathbf{P}_\mathbf{x}$ denotes the covariance matrix of the related states. $r_{m}$ should subject to a $\boldsymbol{\chi}^2$ ``chi-squared'' distribution and thus $r({}^{L_{l+1}} \mathbf{p}_{fi})$ will used in our EKF update if it passes this test.

Similarly, for the projected planar surf features ${}^{L_{l}} \mathbf{p}_{fi}$, we will find  three corresponding surf features, ${}^{L_{l}} \mathbf{p}_{fj}, {}^{L_{l}} \mathbf{p}_{fk}, {}^{L_{l}} \mathbf{p}_{fl}$, which are assumed to be sampled on the same physical plane as ${}^{L_{l}} \mathbf{p}_{fi}$.
The measurement residual of surf feature ${}^{L_{l+1}} \mathbf{p}_{fi}$ is the distance between its projected feature point ${}^{L_{l}} \mathbf{p}_{fi}$ and the plane formed by ${}^{L_{l}} \mathbf{p}_{fj}, {}^{L_{l}} \mathbf{p}_{fk}, {}^{L_{l}} \mathbf{p}_{fl} $. 
The covariance propagation of the distance measurement of surf features, linearization and Mahalanobis distance test are similar to the edge feature.

\subsubsection{Visual Feature Measurement}
Given a new image, we similarly propagate and augment the state.
FAST features are extracted from the image and tracked into future frames using KLT optical flow.
Once a visual feature is lost or has been tracked over the entire sliding window, we triangulate the feature in 3D space using the current estimate of the camera clones \cite{Mourikis2007ICRA}.
The standard visual feature reprojection error is used in the update.
For a given set of feature bearing measurements $\mathbf{z}_i$ of a 3D visual feature ${}^{G}\mathbf{p}_{fi}$ the general linearized residual is:
\begin{align}
	\mathbf{r}(\mathbf{z}_i) &=  \mathbf h(\mathbf{x},{}^{G}\mathbf{p}_{fi}) + \mathbf{n}_r \\
	&= \mathbf h(\hat{\mathbf{x}}, {}^{G}\hat{\mathbf{p}}_{fi}) + \mathbf{H}_\mathbf{x} \delta\mathbf{x} + \mathbf{H}_\mathbf{f} {}^{G}\delta{\mathbf{p}}_{fi} + \mathbf{n}_r
	\label{eq: visual residual}
\end{align}
%
where $\mathbf{H}_f$ is the Jacobian of visual feature measurement with respect to the 3D feature ${}^{G}\mathbf{p}_{fi}$ and both Jacobians are evaluated at the current best estimates.
Since our measurements are a function of ${}^{G}\hat{\mathbf{p}}_{fi}$ (see~\eqref{eq: visual residual}),
we leverage the MSCKF nullspace projection to remove this dependency \cite{Mourikis2007ICRA}.
After the nullspace projection we have:
\begin{align}
	\mathbf{r}_{o}(\mathbf{z}_i) = \mathbf{H}_\mathbf{xo} \delta\mathbf{x} + \mathbf{n}_{ro}
	\label{eq: visual residual_o}
\end{align}
It should be noted that the Jacobian with respect to the rigid transformation between IMU and camera $\{ {}^{C}_I\bar{q}, {}^C\mathbf{p}_{I} \}$ is non-zero, which means the transformation between IMU and camera can be calibrated online.

\subsection{Measurement Compression}

After linearizing the LiDAR feature and visual feature measurements at current state estimate, we could naively perform an EKF update, but this comes with a large computational cost due to the large number of visual and LiDAR feature measurements.
Consider the stack of all measurement residuals and Jacobians (which are from LiDAR  or visual features):
\begin{align}
	\mathbf{r} = \mathbf{H}_\mathbf{x} \delta\mathbf{x} + \mathbf{n}
\end{align}
where $\mathbf r$ and $\mathbf n$ are vectors with block elements of residual and noise in~\eqref{eq: lidar residual} or~\eqref{eq: visual residual_o}.
By commonly assuming all measurements statistically independent, the noise vector $\mathbf{n}$ would be uncorrelated.
To reduce the computational complexity, we employ Givens rotation \cite{golub2012matrix} to perform thin QR 
to compress the measurements~\cite{Mourikis2007ICRA}, i.e., 
\begin{align}
	\mathbf H_{\mathbf x} = \begin{bmatrix} \mathbf Q_{H1} &  \mathbf Q_{H2}\end{bmatrix} \begin{bmatrix} \mathbf T_{H} \\  \mathbf 0\end{bmatrix} 
\end{align}
where $\mathbf Q_{H1}$ and $\mathbf Q_{H2}$ are unitary matrices.
After the measurement compression, we obtain:
\begin{align}
	\mathbf{r}_c = \mathbf{T}_{H}\delta \mathbf{x} + \mathbf{n}_c
\end{align}
where the compressed  Jacobian matrix $\mathbf{T}_{H}$ should be square with  the dimension of the state vector $\mathbf{x}$,
and the compressed noise is given by $\mathbf n_c = \mathbf Q_{H1}^\top \mathbf n$.
This compressed linear measurement residual is then used to efficiently update the state estimate and coviance with the standard EKF.

\section{Experimental Results}

\begin{figure}
	\centering
	\includegraphics[width=0.5\columnwidth]{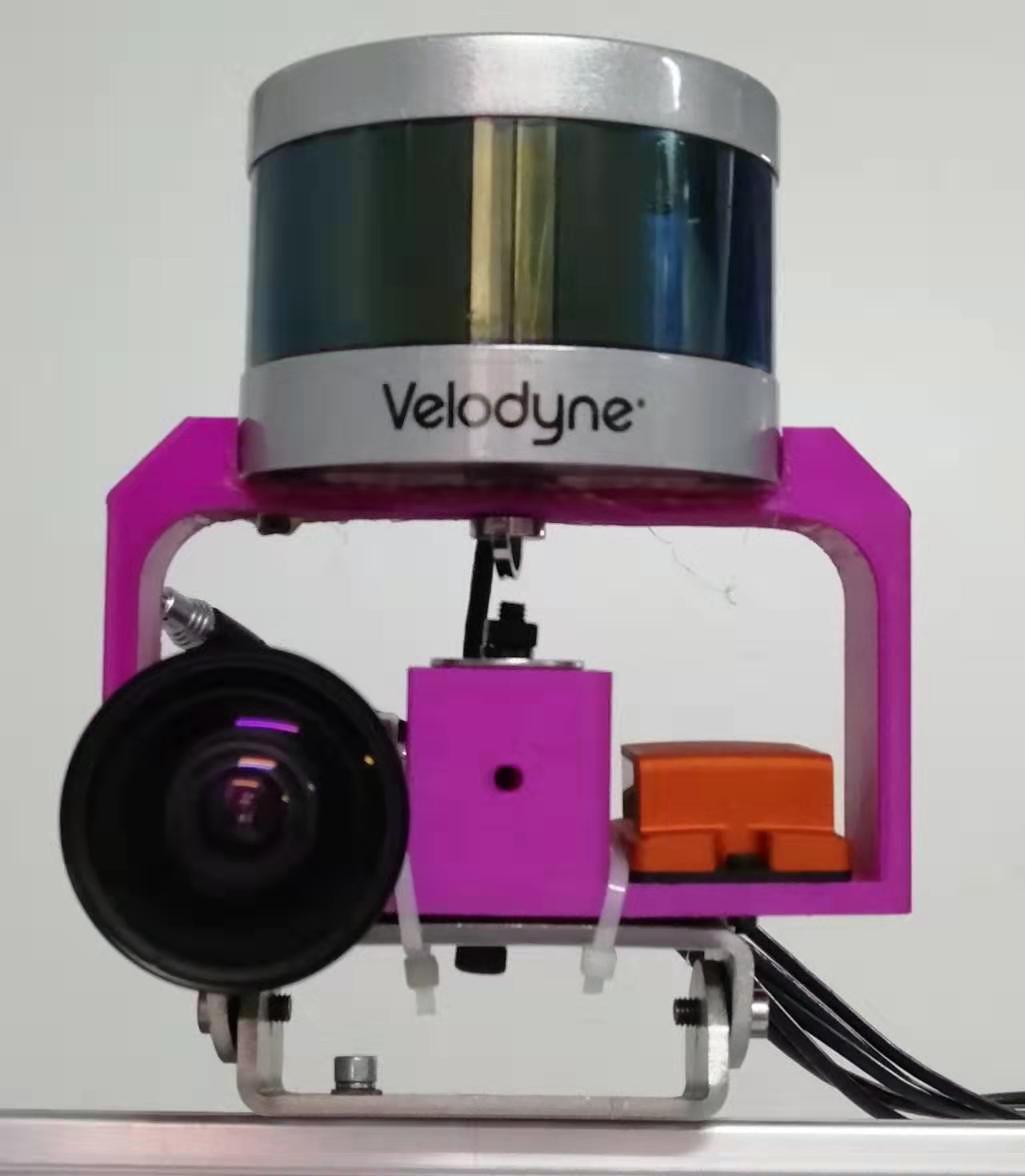}
	\caption{The self-assembled LiDAR-inertial-camera rig with Velodyne LiDAR, Xsens IMU, and monochrome camera.}
	\label{fig:rig}
\end{figure}

To validate the performance of the proposed algorithm, several experiments were performed both in outdoor and indoor environments.
%
The sensor rig, shown in Figure~\ref{fig:rig}, consists of an Xsens MTi-300 AHRS IMU, Velodyne VLP-16 LiDAR, and monochrome global-shutter Blackfly BFLY-PGE-23S6M camera.
The extrinsics between sensors are calibrated offline and refined during online estimation.
For evaluation, we compare the proposed LIC-Fusion against the state-of-the-art visual-inertial and LiDAR odometry methods.
Specifically, we compare the proposed to our implementation of the standard MSCKF-based VIO \cite{Mourikis2007ICRA} and the open sourced implementation of LOAM LiDAR odometry \cite{zhang2014loam}.
It is also important to note that we directly compare to the output of LOAM which leverages ICP matching to its constructed \textit{global map} and thus has leveraged implicit loop-closure information, while our LIC-Fusion is purely an odometry based method which estimates states in a sliding window, neither maintain a global map, nor leverage loop closures.

\subsection{Outdoor Tests}

Firstly, the proposed system is tested on an outdoor sequence collected by mounting the self-assembled sensor rig (see Figure~\ref{fig:rig}) on a custom Ackermann robot platform.
This outdoor sequence is around 800 meters in length and is recorded over a duration of 4 minutes.
RTK GPS with centimeter-level accuracy is also mounted on and the GPS measurements are used as the groundtruth for evaluation.
%
%
\begin{table}[]
	\centering
	\caption{
		Outdoor Experimental Results: Average of average absolute trajectory errors (ATE) and their standard deviation/variability.
	}
	
	\begin{tabular}{C{2.25cm}C{1.5cm}C{1.5cm}C{1.5cm}} \toprule
		& \textbf{MSCKF \cite{Mourikis2007ICRA}} & \textbf{LIC-Fusion} & \textbf{LOAM \cite{zhang2014loam}} \\ \hline
		{Average ATEs (m)}  & {10.75}  & {4.06}     & {23.08}      \\ 
		{1 Sigma (m)}  & {3.56}  & {3.42}     & {2.63}      \\ 
		\hline
	\end{tabular}
	\color{black}
	\label{table:result1}
\end{table}

\begin{figure}[t]
	\centering
	\includegraphics[width=0.85\columnwidth]{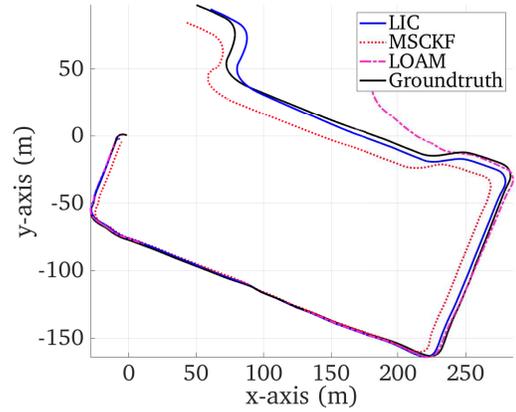}\vspace{-2em}
	\caption{Top view of outdoor sequence trajectories, showing the trajectories resulted from proposed LIC-Fusion (blue), MSCKF (red), LOAM (pink), and RTK GPS groundtruth (black)}
	\label{fig:GPS1_xy_plot}
\end{figure}
\begin{figure}[th]
	\includegraphics[width=0.95\columnwidth] {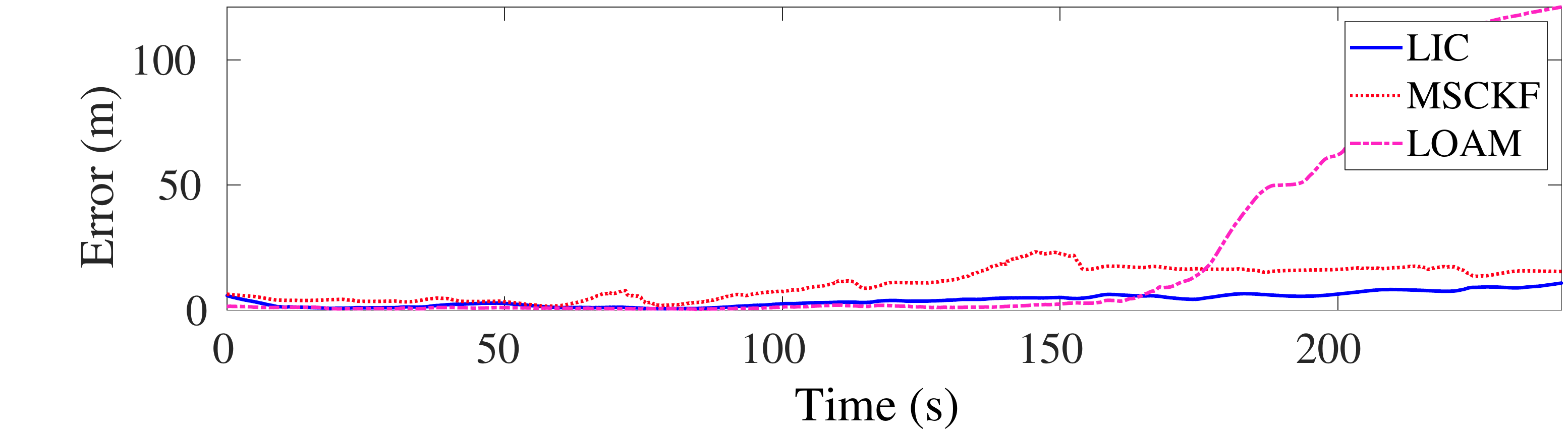}
	\caption{Average mean squared errors (MSE) of the proposed LIC-Fusion (blue), MSCKF (red), and LOAM (pink) on the outdoor sequence, over the duration of the trajectory.}
	\label{fig:norm}
\end{figure}

%
%
%
Each algorithm was run six different times to account for their inherent randomness due to the use of RANSAC and to provide a representative evaluation of typical performance.
Figure \ref{fig:GPS1_xy_plot} shows the resulting mean trajectories estimated by the proposed LIC-Fusion, MSCKF, and LOAM.
The average mean squared errors (MSE) of each method is presented in Fig. \ref{fig:norm}, in which the trajectories are aligned to the RTK groundtruth using the ``best fit'' transform that minimized the overall trajectory error.
The proposed LIC-Fusion showed a 2.5 meter decrease in the average error as compared to the standard MSCKF, and 5 meter decreased when compared to LOAM.
We can find that the drift of LIC-Fusion grows much slower over time as compared to the other two methods and maintains the smallest error for most of the trajectory.
The average absolute trajectory errors (ATE)~\cite{sturm2012benchmark} and their one sigma deviation/variability are also reported in Table \ref{table:result1}.
These results show that the proposed system is able to localize with high accuracy by fusing different sensing modalities (that being camera, inertial, and LiDAR).

\begin{figure*}
	\centering
	\begin{subfigure}{.23\textwidth}
		\centering
		\includegraphics[width=\linewidth]{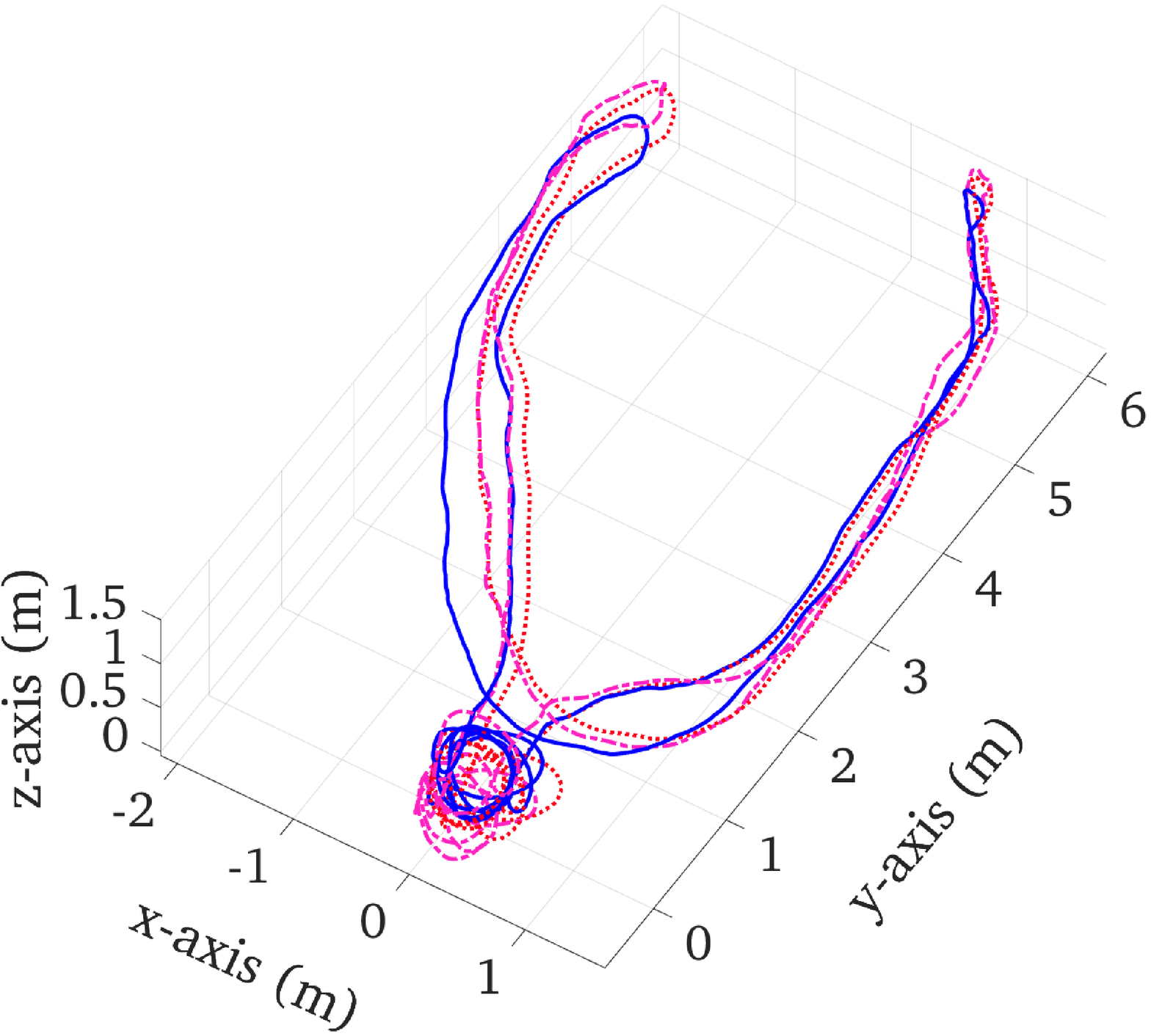} %
		\label{fig:indoorA}
	\end{subfigure}
	\begin{subfigure}{.23\textwidth}
		\centering
		\includegraphics[width=\linewidth]{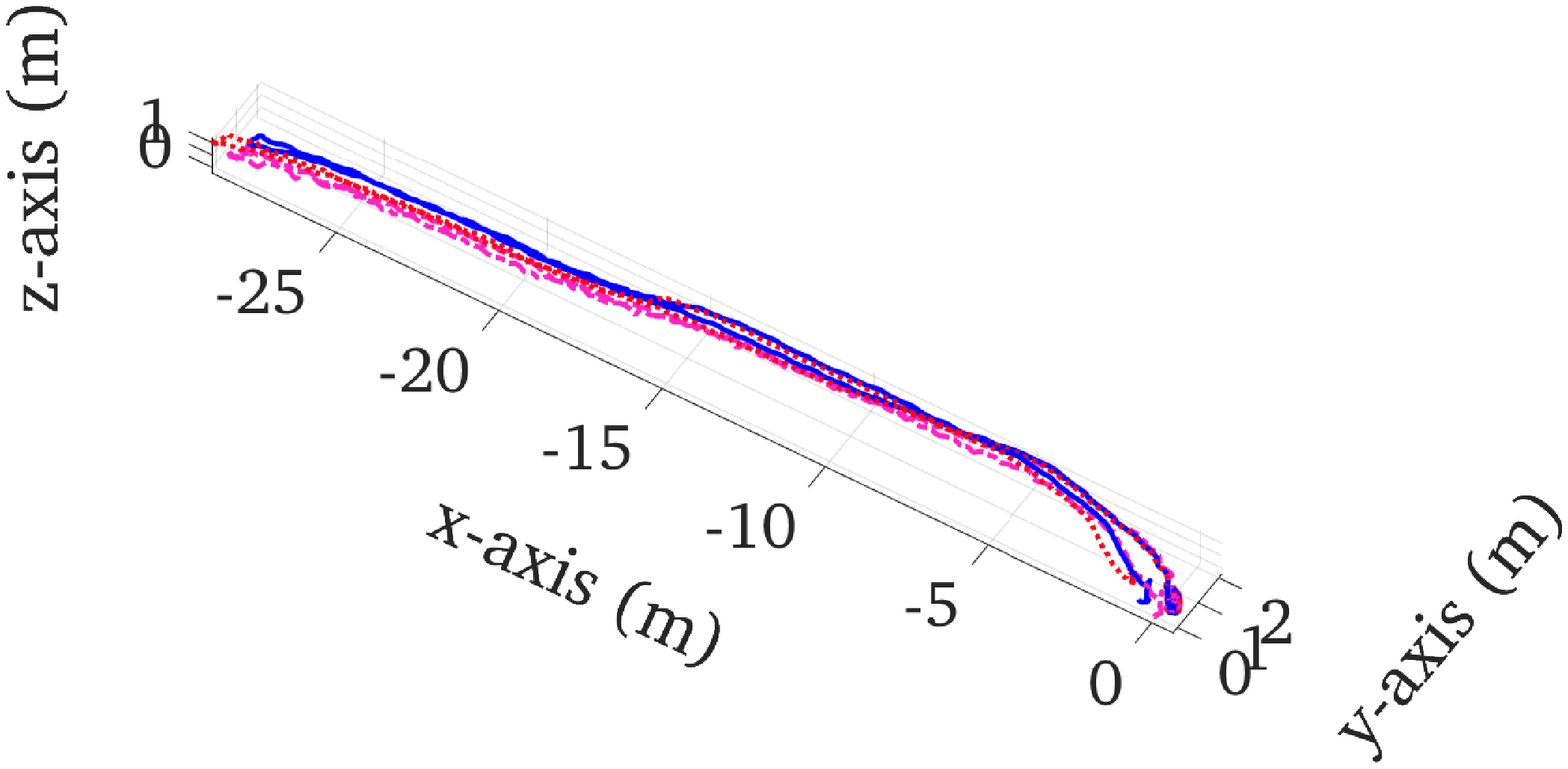} %
		\label{fig:indoorB}
	\end{subfigure}
	\begin{subfigure}{.23\textwidth}
		\centering
		\includegraphics[width=\linewidth]{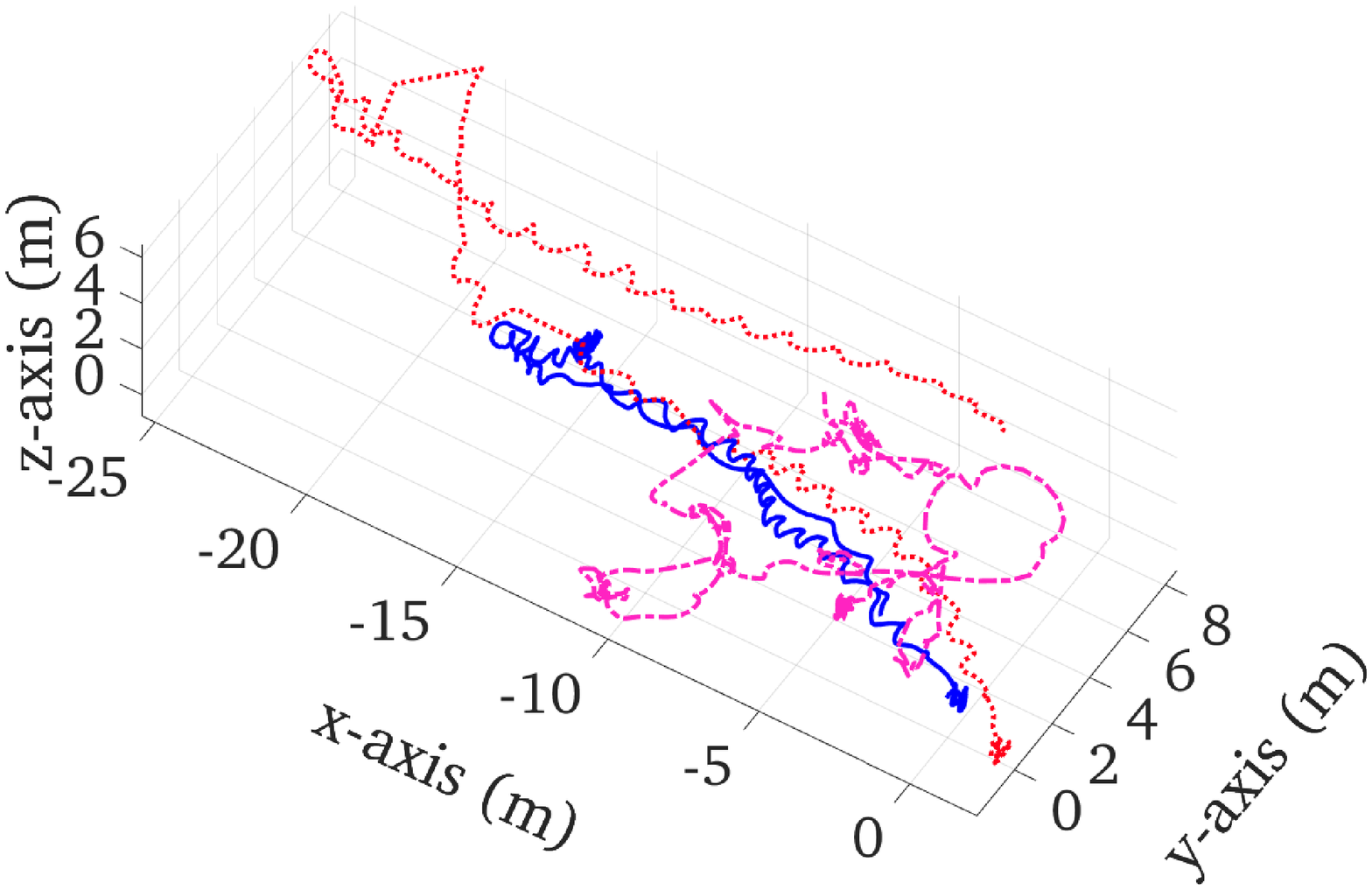} %
		\label{fig:indoorC}
	\end{subfigure}
	\begin{subfigure}{.23\textwidth}
		\centering
		\includegraphics[width=\linewidth]{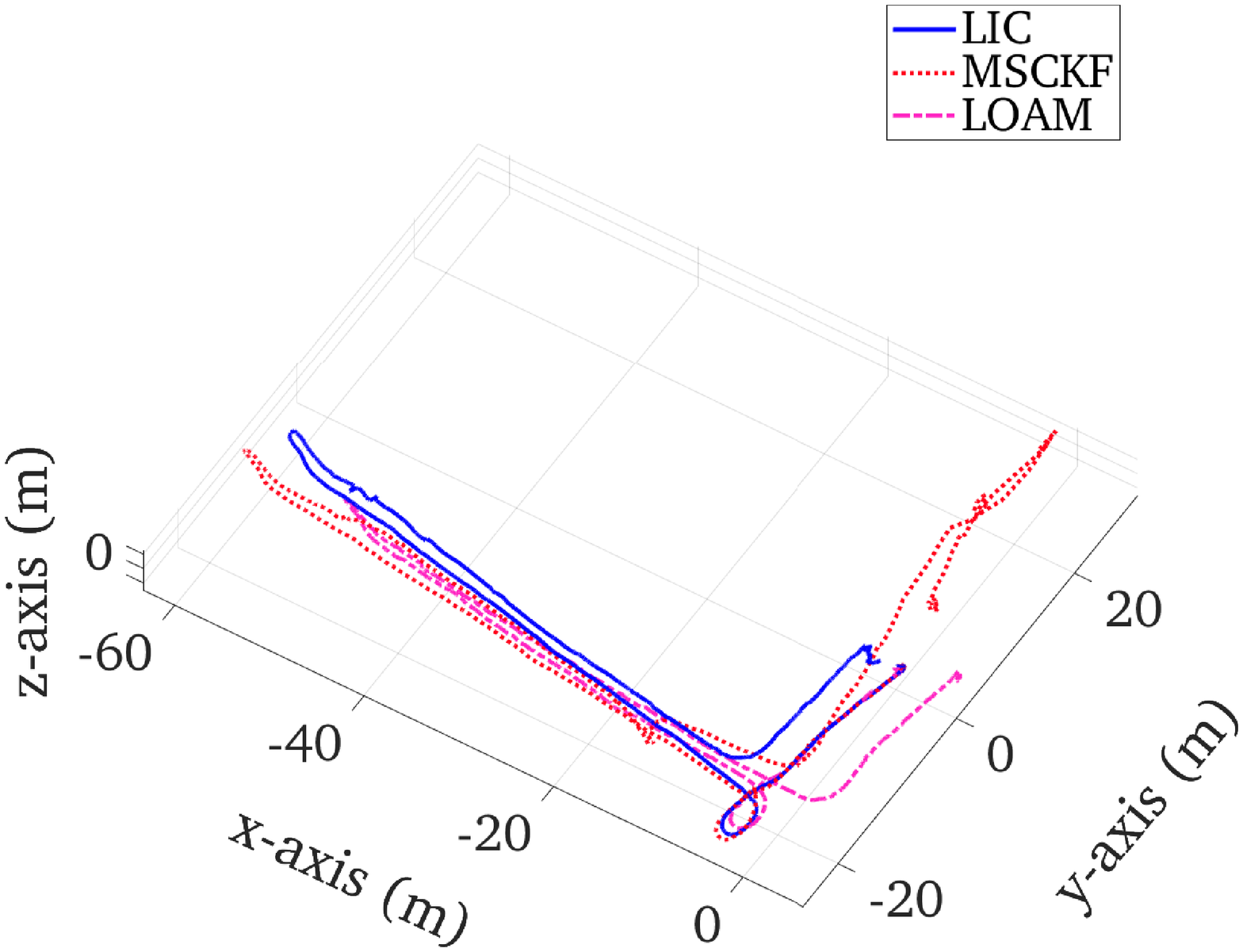} %
		\label{fig:indoorD}
	\end{subfigure}
	\caption{Isometric views of the estimated trajectories on indoor sequences A, B, C and D (from left to right).}%
	\label{fig:indoortraj}
\end{figure*}

\subsection{Indoor Tests}

\begin{figure}
	\centering
	\includegraphics[width=\columnwidth] {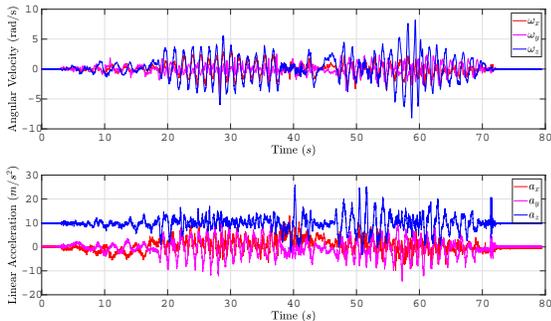}
	\caption{Raw IMU measurements over the high-dynamic Indoor-C sequence.}
	\label{fig:indoor_c}
\end{figure}

%

We further evaluate the system on a series of indoor datasets which were collected in various normal to low-light lighting conditions with slow to aggressive motion profiles.
The indoor sequences are collected by holding the sensor rig (see Figure~\ref{fig:rig}) in hand at chest height.
Since groundtruth was not available indoors, we returned the sensor platform to the initial location and evaluate the start-end error.
Table \ref{table:result2}, summarizes the average start-end error results with the trajectories being shown in Figure \ref{fig:indoortraj}.
The results show that the proposed LIC-Fusion is able to localize with high accuracy and is able to handle even extreme cases of high motion and low light due to the fusion of three different sensing modalities.
Shown in Figure \ref{fig:indoor_c}, the Indoor-C sequence recorded while we shook the sensor rig as strongly as we could, hence it has both high angular velocities and high linear accelerations with aggressive motion.
The proposed LIC-Fusion is able to localize in this sequence, while the compared two methods fail with large amounts of errors.

\begin{table}[t]
	\centering
	\caption{
		Indoor Experimental Results: Average trajectory start-end errors 
	}
	\begin{tabular}{C{2.0cm}C{1.5cm}C{1.5cm}C{1.5cm}} \toprule
		\textbf{Sequence} & \textbf{MSCKF \cite{Mourikis2007ICRA}} & \textbf{LIC-Fusion} & \textbf{LOAM \cite{zhang2014loam}} \\ \hline
		{Indoor-A (39m)}  & {0.99}  & {0.98}     & {0.66}      \\ 
		{Indoor-B (86m)}  & {1.55}  & {1.04}     & {0.46}      \\ 
		{Indoor-C (55m)}  & {49.94} & {1.55}     & {2.44}      \\ 
		{Indoor-D (189m)} & {46.03} & {3.68}     & {5.99}      \\
		\hline
	\end{tabular}
	\color{black}
	\label{table:result2}
\end{table}

\section{Conclusions and Future Work}

In this paper, we have developed a tightly-coupled efficient multi-modal sensor fusion algorithm for LiDAR-inertial-camera odometry (i.e., LIC-Fusion) within the MSCKF framework.
Online spatial and temporal calibration between all three sensors is  performed to compensate for calibration sensitivities as well as  to ease sensor deployment.
The proposed approach detects and tracks sparse edge and planar surf feature points over LiDAR scans and fuses these measurements along with the visual features extracted from monocular images.
As a result, by taking advantages of different sensing modalities, the proposed LIC-Fusion  is able to provide accurate and robust 6DOF motion tracking in 3D in different environments and under aggressive motions.
In the future, we will investigate how to efficiently integrate loop closure constraints obtained from both the LiDAR and camera in order to bound navigation errors.


{
 \vspace{0.05cm}

\bibliographystyle{ieeetr}
\bibliography{main}

}

\end{document}